\documentclass[letterpaper, 10 pt, conference]{ieeeconf}  

\IEEEoverridecommandlockouts                              

\title{\LARGE \bf
Fast LiDAR Upsampling using Conditional Diffusion Models
}

\author{Sander Elias Magnussen Helgesen$^{1}$, 
        Kazuto Nakashima$^{2}$, 
        Jim Tørresen$^{1}$ 
        and Ryo Kurazume$^{2}$
\thanks{*This work was partially supported by JSPS KAKENHI Grant Number JP23K16974 and JP20H00230, The Research Council of Norway (RCN) as a part of the projects: Collaboration on Intelligent Machines (COINMAC) project, under grant agreement no. 309869, Vulnerability in the Robot Society (VIROS) under Grant Agreement No. 288285, Predictive and Intuitive Robot Companion (PIRC) under grant agreement no. 312333 and through its Centres of Excellence scheme, RITMO with project no. 262762.}
\thanks{$^{1}$Sander Elias Magnussen Helgesen and Jim Tørresen are with the Department of Informatics,
        University of Oslo, Norway.
        {\tt\small \{sehelges, jimtoer\}@ifi.uio.no}}%
\thanks{$^{2}$Kazuto Nakashima and Ryo Kurazume are with the Faculty of Information Science and Electrical Engineering,
        Kyushu University, Japan.
        {\tt\small \{k\_nakashima, kurazume\}@ait.kyushu-u.ac.jp}}%
}

\usepackage{xcolor}
\usepackage{float}
\usepackage{graphicx}
\usepackage{multirow}
\usepackage{longtable}
\usepackage{threeparttable}
\usepackage{wrapfig}
\usepackage{booktabs}
\usepackage{tabularx}
\usepackage[space]{cite}
\usepackage{amssymb}
\usepackage{ulem}

\newcolumntype{C}{>{\centering\arraybackslash}X}
\newcolumntype{L}{>{\raggedright\arraybackslash}X}
\newcolumntype{R}{>{\raggedleft\arraybackslash}X}
\newcommand{\etal}{\textit{et al.} }

\graphicspath{images/}

\begin{document}

\maketitle
\thispagestyle{empty}
\pagestyle{empty}

\begin{abstract}

The search for refining 3D LiDAR data has attracted growing interest motivated by recent techniques such as supervised learning or generative model-based methods. Existing approaches have shown the possibilities for using diffusion models to generate refined LiDAR data with high fidelity, although the performance and speed of such methods have been limited. These limitations make it difficult to execute in real-time, causing the approaches to struggle in real-world tasks such as autonomous navigation and human-robot interaction. In this work, we introduce a novel approach based on conditional diffusion models for fast and high-quality sparse-to-dense upsampling of 3D scene point clouds through an image representation. Our method employs denoising diffusion probabilistic models trained with conditional inpainting masks, which have been shown to give high performance on image completion tasks. We introduce a series of experiments, including multiple datasets, sampling steps, and conditional masks. This paper illustrates that our method outperforms the baselines in sampling speed and quality on upsampling tasks using the KITTI-360 dataset. Furthermore, we illustrate the generalization ability of our approach by simultaneously training on real-world and synthetic datasets, introducing variance in quality and environments. 

\end{abstract}

\section{Introduction}

3D LiDAR (Light Detection And Ranging) plays a great role in mobile robotics, such as self-driving cars, humanoids, personal mobility devices, and service robots interacting with humans.
It enables the robots to perceive surrounding obstacles and geometry for mapping, autonomous navigation, and safer and better interaction with humans, such as assisting an elderly person in their home.

However, due to hardware constraints and the costs of LiDAR sensors, the quality and density of measurement data vary significantly.
In particular, the vertical resolution of LiDAR sensors, defined by the number of emitted beams, varies widely, ranging from 8 to 64, for example.
More beams allow for capturing more information from the scans.
This variability could lead to techniques such as semantic segmentation~\cite{milioto2019rangenet} and object detection to suffer from inconsistent performance, which is suboptimal for operational robots.

To transform data between the sensors with different quality and beam resolution, previous studies have suggested the use of image-based LiDAR data generation using deep generative models such as generative adversarial networks (GANs)~\cite{nakashima2023generative}. However, these techniques encounter issues with stability during training and the fidelity of generated data.
In recent studies~\cite{zyrianov2022learning, nakashima2024lidar}, diffusion models~\cite{song2020improved,ho2020denoising} have been proposed as an alternative, improving the diversity and fidelity of generated 3D scene point clouds.
The studies leveraged the pre-trained unconditional diffusion models to solve an upsampling task, however, the approaches involve complex procedures that result in slow inference times, making them unsuitable for real-time robot navigation pipelines.

\begin{figure}[t]
    \centering
    \includegraphics[width=\hsize]{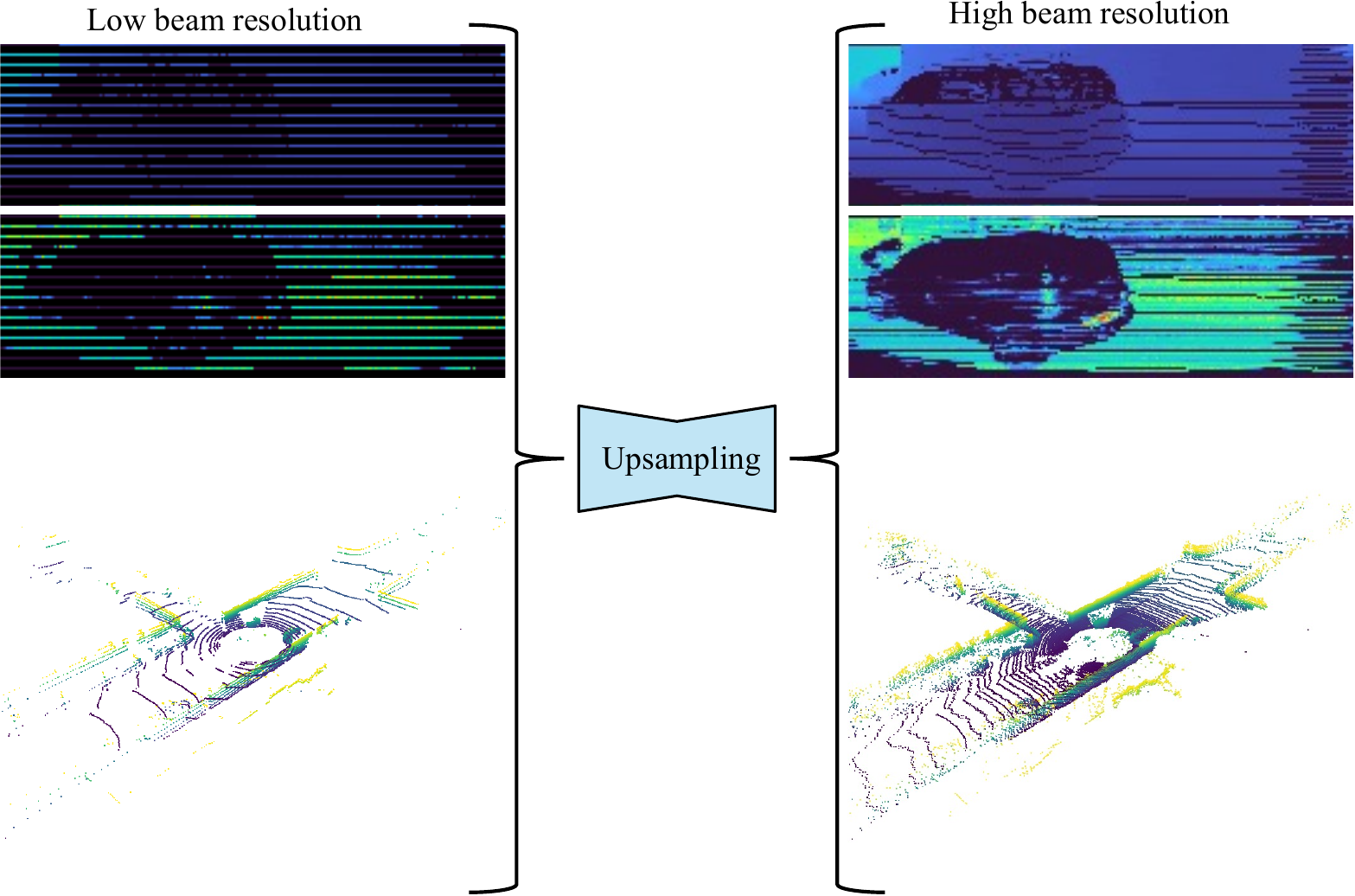}
    \caption[Upsampling example]{\textbf{Upsampling results using our method.} We show the row-resolution input (left) and the high-resolution output obtained by our conditional diffusion model (right), each with the range image (top), the reflectance image (middle), and the point cloud converted from the range image (bottom ). Compared to the existing approaches with 1,160 and 320 sampling steps~\cite{zyrianov2022learning,nakashima2024lidar}, our method can produce better results in only 8 steps. For visual purposes, the images for reflectance and range have been cropped from the original $64 \times 1024$ images.}
    \label{upsampling_example}
\end{figure}

This paper aims to speed up the diffusion-based LiDAR upsampling method, maintaining the high fidelity demonstrated in the previous work.
To this end, we build a conditional diffusion model~\cite{saharia2022palette} that learns the LiDAR data generation given the partial observation.
The conditional diffusion models eliminate the objective gap between training and sampling.
Moreover, the training task gets easier, generating only a partial extent by conditioning with observed context.
For stability in training, high sample quality, and great generalizability, we base our architecture on denoising diffusion probabilistic models (DDPMs)~\cite{ho2020denoising,nakashima2024lidar}. Fig. \ref{upsampling_example} shows an example of $4\times$ upsampling of lidar data using our approach.

As suggested in relevant studies~\cite{zyrianov2022learning, nakashima2024lidar, caccia2019deep}, we represent 3D LiDAR point clouds as equirectangular images in which each pixel contains the according point range and reflectance information, and perform the LiDAR upsampling as an image inpainting. Motivated by the natural image domain~\cite{saharia2022palette}, we experiment with combining multiple inpainting tasks for training our conditional diffusion model.
In addition, we experiment with combining datasets of various environments to show the robustness and generalizability of our technique. We show that our proposed technique achieves state-of-the-art performance on the upsampling task. 
The following summarizes our contributions:

\begin{itemize}
    \item We present a fast and effective LiDAR upsampling method combined with denoising diffusion probabilistic models (DDPMs). Our approach shows improved performance with significantly faster inference compared to existing baselines on the KITTI-360 \cite{liao2022kitti} dataset.
    \item We explored the training task design for our conditional diffusion model, including various training masks and multiple datasets.
\end{itemize}

\section{Related Work}

\subsection{Image inpainting using diffusion models}

Diffusion-based models are a branch within the field of generative models. They aim to learn the general distribution of a set of data. More specifically a branch of diffusion-based models that has gained much traction and has been shown to learn and generate high-quality samples effectively, is called denoising diffusion probabilistic models (DPPMs)~\cite{ho2020denoising}. It works by iteratively adding noise to the input data and learning the general distribution to denoise the image. 

DDPMs can be used for various tasks such as generating similar samples to a dataset and completing tasks where DDPMs have learned to perform tasks such as denoising, upsampling, upscaling, and many more. Saharia \etal introduced the Palette model~\cite{saharia2022palette}. The Palette architecture was trained on 4 separate tasks: colorization, inpainting, uncropping, and JPEG restoration. The model was shown to give improved overall performance when trained on all tasks combined (multi-task diffusion) compared to task-specific trained models. The architecture works by giving a condition under training, meaning that extra information or data is fed in as a set with the input data. In the case of the inpainting task, each image is masked with a free-form generated mask that removes a section of the image, this masked area is then filled with Gaussian noise which allows the DDPM to learn to fill in the masked area. This allows for a speed-up during training as the loss only considers the images' masked area. 

Another well-established completion technique is the RePaint technique~\cite{lugmayr2022repaint}. RePaint is a technique that builds on the unconditionally trained DDPMs on the task of inpainting. Differently from the Palette inpainting technique, RePaint is conditioned by sampling on a given mask during the iterations of the reverse diffusion process and not during the initial forward steps. As the pixels are not initialized with noise like the inpainting technique, RePaint works by repeatedly closing the gap between the known and unknown pixels through the reverse and forward process of the DDPM~\cite{lugmayr2022repaint, nakashima2024lidar}. 

\subsection{LiDAR applications}

As a result of great success within the domain of images and recent advancements in generative models for LiDAR, several studies have shown generative models to be applicable also in the domain of inpainting and completion tasks with LiDAR data. Nakashima and Kurazume~\cite{nakashima2024lidar} introduced the R2DM, a pipeline that trains a DDPM for \textit{unconditional} generation and upsamples LiDAR data with a post-hoc resampling technique~\cite{lugmayr2022repaint}. Nakashima and Kurazume~\cite{nakashima2024lidar} further showed that the geometrical consistency of generated samples was greatly influenced by introducing explicit spatial bias to the architecture. 

R2DM was compared against the baseline LiDARGen~\cite{zyrianov2022learning} and the non-diffusion model-based GAN-based methods~\cite{caccia2019deep, nakashima2021learning, nakashima2023generative}. It was tested against the baseline LiDARGen on generation and upsampling on the KITTI-360 dataset~\cite{liao2022kitti} with resolution $64 \times 1024$, and against non-diffusion model-based methods it was tested on the KITTI-Raw dataset~\cite{geiger2013vision} with $64 \times 512$ subsampling from point clouds projected on the full resolution of $64 \times 2048$. R2DM was shown to outperform LiDARGen on tested metrics and had a great reduction in sampling steps required to obtain the same or better quality and results.

Kwon \etal proposed Implicit LiDAR Network (ILN)~\cite{kwon2022implicit} for LiDAR super-resolution. ILN learns the neighbor pixel interpolation weights so that super-resolution can be performed by blending input pixel depths with non-linear weights. The connection between neighbor pixels is further seen as attention similar to the attention module in Transformer~\cite{vaswani2017attention} models. 

ILN was trained and tested on several resolutions to the up-scaling tasks. The input resolution of $16\times 1024$ is up-scaled to three test resolutions: $ 64\times 1024$, $ 128\times 2048$ and $ 256\times 4096$. In each test, ILN was shown to outperform the baselines of Bilinear interpolation, LiDAR-SR~\cite{shan2020simulation}, and LIIF~\cite{chen2021learning} on several metrics.

\begin{figure*}[ht]
    \includegraphics[width=0.95\hsize]{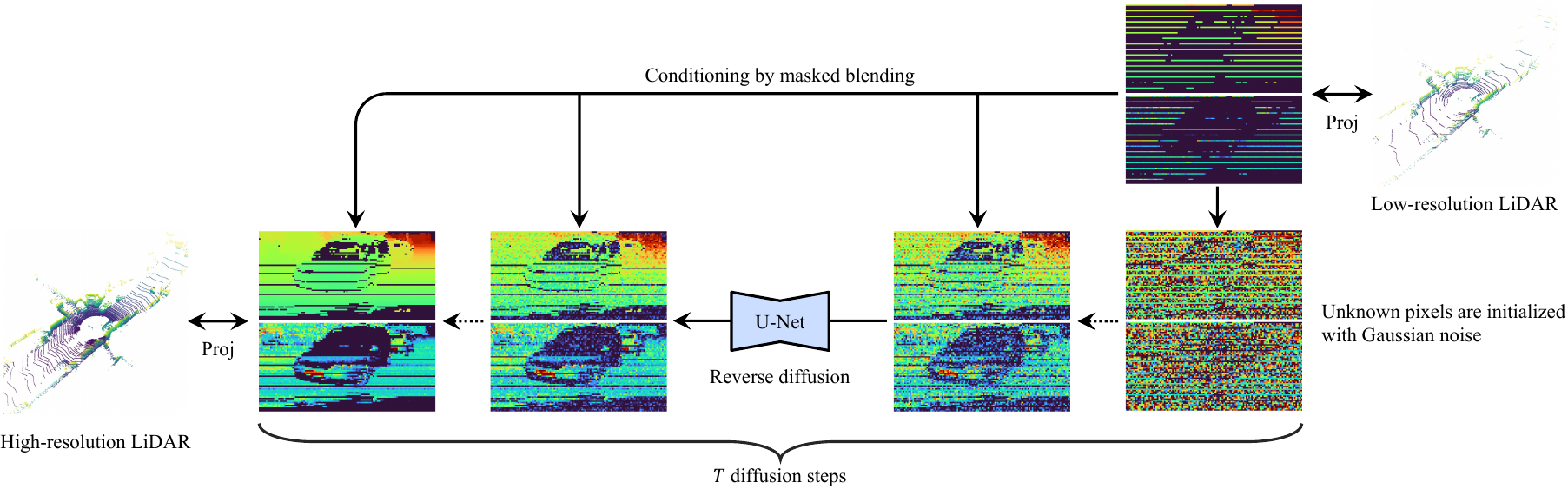}
    \caption{Overview of our upsampling method using a diffusion model. In each reverse diffusion step, the known region is re-initialized by masked blending. \textit{Proj} mean translation between point clouds and images using spherical projection.}
    \label{fig:overview}
\end{figure*}

\section{Proposed Method}

This section introduces data representation, an inpainting method, a model architecture, and a variety of inpainting masks for training.
Fig.~\ref{fig:overview} shows the overview of our method.

\subsection{Data representation} 

Spherical projection is used to transform the 3D point clouds down to a 2D image. This is done using azimuth and elevation angles to project the range distance $d$ onto $H \times W$ angular grid that can be processed as an $H \times W$ image.

For preprocessing, the depth image is scaled using a logarithmic normalization~\cite{zyrianov2022learning, nakashima2024lidar}

\begin{equation}
    d_{log} = \frac{\log(d+1)}{\log(d_{max} + 1)}
    \label{log_normalization}
\end{equation}

The normalization of the intensity image is done using a min-max normalization setting the min intensity to 0 and max intensity to 1. 

\subsection{Image-based inpainting}

In this work, we build an upsampling pipeline for 3D LiDAR, by integrating the DDPM architecture of LiDAR data generation~\cite{nakashima2024lidar} and the inpainting framework inspired by Palette~\cite{saharia2022palette}. In the standard DDPMs, the data generation is learned as an iterative denoising process from latent variables $z$ (Gaussian noise) into data $x$, called reverse diffusion.
Each denoising step is performed by a neural network such as U-Net.
In contrast, our approach combines conditional observable pixels $\hat{x}$ into the input $z_t$ of the neural network for each reverse diffusion step $t$, as depicted in Fig.~\ref{fig:overview}. The observable pixels $\hat{x}$ are blended into the latent variables $z_t$ according to the mask $m$ representing known and unknown region: $m \odot \hat{x} + (1-m) \odot z_t$. Although the upsampling technique~\cite{lugmayr2022repaint} used in R2DM~\cite{nakashima2024lidar} also uses masked sampling, extra harmonization steps are required due to the objective gap between the training and sampling.
For the loss function, we modify the one used in the R2DM~\cite{nakashima2024lidar}, by making the loss only be calculated on the predicted masked area and not the entire image. 

\subsection{Model architecture}
The pipeline's core is the neural network that learns the distribution of the data. We use the Efficient U-net architecture~\cite{saharia2022photorealistic} for the base of our DDPM pipeline. It is shown to have increased performance and efficiency with higher convergence speed, by increasing the residual blocks in the lower-resolution layers and reversing the order of the convolution and sampling blocks shifting convolution before upsampling and after downsampling. 

\subsection{Training method}
Similar to Nakashima and Kurazume~\cite{nakashima2024lidar} we built an upsampling pipeline that takes in a LiDAR point cloud projected to an image and upsamples the input to a $64 \times 1024$ image. In addition, we added 4 inpainting masks for multitask-learning as suggested by Saharia \etal~\cite{saharia2022palette}. Saharia \etal showed that training Palette on several tasks made the model generalize better and perform as well if not better on several tasks. 

\subsubsection{Random straight lines}
The random straight lines mask masks an adjustable ratio of rows in the input image. 

\subsubsection{Random jitter lines}
This mask follows a random pattern along the rows of the input image. The ratio of lines in the mask can be adjusted. 

\subsubsection{Random points (Pepper noise)}
The random points mask is created as a randomly distributed amount of points in the input image that gets masked.

\subsubsection{Even-interval straight lines (Upsampling)}
In the upsampling mask, rows are systematically selected in the ratio selected. If 4x upsampling is selected, all but every fourth row will be included in the mask leaving $1/4$ of the original input image, creating a downsampled image.

\begin{figure}[!t]
    \centering
    \includegraphics[width=\hsize]{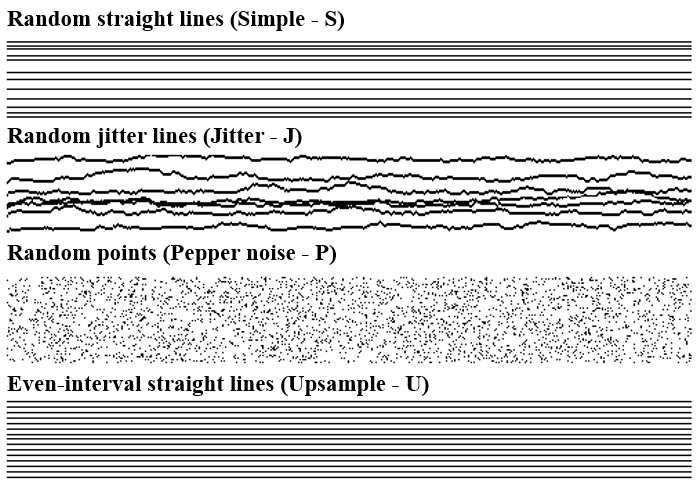}
    \caption[Inpainting masks example]{All masks are originally generated for $64 \times 1024$ images, but for readability, the following examples have been resized to $64 \times 512$. Each mask is binary and the black area is the area affected by the mask.}
    
    \label{Inpainting_simple}
\end{figure}

For simplicity, Fig.~\ref{Inpainting_simple} illustrates how each mask is applied to each input image. The black area is the masked area, while the remaining white area is the original image not affected by the mask.

\section{Experiments}

\subsection{Settings}

To evaluate and measure the performance of our model, we test our models on a 4x upsampling using the upsampling mask from Fig. \ref{Inpainting_simple}. 
We base our experiments on the KITTI-360 dataset~\cite{liao2022kitti}, where samples are projected as $64 \times 1024$ images. 
The KITTI-360 dataset consists of 81,106 samples representing point clouds measured using the Velodyne HDL-64E sensor.
We follow the same split for the data splits as the baselines~\cite{zyrianov2022learning,nakashima2024lidar}.
Additionally, we use the KITTI-Raw~\cite{geiger2013vision} and SynLiDAR~\cite{xiao2022transfer} datasets to augment the training data. 

In addition to the R2DM baseline, we also compare our methods against well-established interpolation techniques (Nearest-neighbor, Bilinear, and Bicubic). We also compare our results against the LiDAR diffusion model based LiDARGen~\cite{zyrianov2022learning} on 4x upsampling. Lastly, we compare with simulation-based supervised completion methods: Simulation base LiDAR super-resolution (LiDAR-SR)~\cite{shan2020simulation}, and Implicit LiDAR Network (ILN)~\cite{kwon2022implicit}.

\subsection{Evaluation metrics}

As a considerable section of each sample keeps the original data, each metric is only calculated for the generated area decided by the mask to measure improvements effectively. 

\subsubsection{Intersection over Union (IoU) score}

Following the setup of Zyrianov \etal~\cite{zyrianov2022learning}, we utilize a pre-trained unmodified RangeNet$++$~\cite{milioto2019rangenet} to calculate the IoU between the labels predicted from the upsampled results and the ground truths.

\subsubsection{Mean Absolute Error (MAE)}

Mean Absolute Error (MAE) is calculated for depth and reflectance image respectively. MAE is calculated by taking the mean of pixel-wise absolute errors between prediction and its ground truth. 

\subsubsection{Root Mean Squared Error (RMSE)}

Root Mean Square Error (RMSE) is calculated by taking the root of the mean difference between predictions and the ground truths.

\subsection{Implementation details}
 
For fair comparison, we use the same parameter setup as the baseline~\cite{nakashima2024lidar}.
Training was performed on two NVIDIA RTX3090 GPUs. The average training of a model took approximately 12 GPU hours and VRAM 7 GB. An estimated 9 minutes to 5.5 GPU hours were spent on inpainting around 30k samples in \{2,4,8,16,32,64,128\} denoising steps per model.

\subsection{Results}

\begin{table*}[t] 
	\centering
	\begin{threeparttable}
		\caption{Experiment on 4x upsampling}
		\label{upsampling_experiments}
		\begin{tabularx}{\hsize}{lcc CCCC lllll}
			\toprule
			& & & \multicolumn{4}{c}{\textbf{Training task$^\dag$}} & \multicolumn{2}{c}{\textbf{Depth}} & \multicolumn{2}{c}{\textbf{Reflectance}} & \\
			\cmidrule(lr){4-7} \cmidrule(lr){8-9} \cmidrule(lr){10-11}
			\textbf{Method}                & \textbf{Timesteps} & \textbf{Train dataset} & \textbf{S} & \textbf{J} & \textbf{P} & \textbf{U} & \textbf{MAE $\downarrow$} & \textbf{RMSE $\downarrow$} & \textbf{MAE $\downarrow$} & \textbf{RMSE $\downarrow$} & \textbf{IoU\% $\uparrow$} \\
			\midrule
			R2DM~\cite{nakashima2024lidar} & 320                & KITTI-360              &            &            &            &            & 0.9229                      & 3.9108                       & 0.0503                      & 0.1052                       & 34.43                   \\
			\midrule
			Ours A                         & 8                  & KITTI-360              &            &            &            & 2-8$\times^\ddag$       & \uline{0.6578}              & 3.4690                       & \uline{0.0407}              & \uline{0.0858}               & \uline{44.27}           \\
			Ours B                         & 8                  & KITTI-360              &            &            &            & 2$\times$         & 8.7424                      & 15.9029                      & 0.1778                      & 0.2652                       & 3.81                    \\
			Ours C                         & 8                  & KITTI-360              &            &            &            & 4$\times$         & \uline{0.6431}              & \uline{3.3912}               & \uline{\textbf{0.0402}}     & \uline{\textbf{0.0844}}      & \uline{\textbf{45.55}}  \\
			Ours D                         & 8                  & KITTI-360              &            &            &            & 8$\times$         & 1.0010                      & 3.9631                       & 0.0554                      & 0.1103                       & 30.58                   \\
			\midrule
			Ours E                         & 8                  & KITTI-360              & \checkmark & \checkmark & \checkmark & 4$\times$         & 0.8016                      & 3.6900                       & 0.0460                      & 0.0934                       & 37.50                   \\
			Ours F                         & 8                  & KITTI-360              & \checkmark & \checkmark & \checkmark & 2-8$\times$       & 0.6868                      & \uline{3.4619}               & 0.0422                      & 0.0880                       & 42.44                   \\
			Ours G                         & 8                  & KITTI-360              & \checkmark &            & \checkmark & 4$\times$         & 0.6845                      & 3.4757                       & \uline{0.0420}              & 0.0876                       & 42.66                   \\
			Ours H                         & 8                  & KITTI-360              & \checkmark & \checkmark &            & 4$\times$         & 0.7472                      & 3.6017                       & 0.0433                      & 0.0900                       & 40.33                   \\
			Ours I                         & 8                  & KITTI-360              &            & \checkmark &            & 4$\times$         & 0.7238                      & 3.5826                       & 0.0433                      & 0.0896                       & 39.64                   \\
			\midrule
			Ours J                         & 8                  & All datasets           &            &            &            & 4$\times$         & \uline{\textbf{0.6299}}     & \uline{\textbf{3.3474}}      & 0.0422                      & \uline{0.0867 }              & \uline{44.58}           \\
			\bottomrule
		\end{tabularx}
		\begin{tablenotes}
                \item The best scores and the top 3 scores are highlighted in bold and underline, respectively.
		      \item \dag To keep the table readable we have abbreviated each mask combination. Each letter is related to each mask: Simple (S), Jitter (J), Pepper noise (P), and Upsampling (U). Under the Upsampling column, we have included the upsampling ratios.
                \item \ddag 2-8x states that samples during training were randomly masked with one of the optional 2x, 4x, and 8x upsampling rates respectively.
		\end{tablenotes}
	\end{threeparttable}
\end{table*}

Table \ref{upsampling_experiments} lists all training configurations (config A-J) for experiments on 4x upsampling. 
Table \ref{upsampling_experiments} contains three types of experiments mainly focusing on the combinations of training masks. Configuration A-D experiment with various upsampling rates, configuration E-I compares the different combinations of the mask types, and configuration J is trained on three datasets (KITTI-360~\cite{liao2022kitti}, KITTI-Raw~\cite{behley2019semantickitti} and SynLiDAR~\cite{xiao2022transfer}).

Although our model demonstrates improvement with each increment of $T$, we will refer to our inpainting DDPM as $T=8$ in the following experiments unless otherwise specified, as it is considered to strike the optimal balance between computational efficiency and performance. We observe a small spike around 4-8 timesteps in Fig. \ref{best_model_compare} in both MAE and RMSE for intensity. This could be partially due to subtle smoothing in smaller details as these metrics are calculated pixel-wise and calculating the mean at the end. Therefore smaller changes in details such as edges and objects could be canceled out due to pixels missing evenly on both sides of the target value, giving a lower score. Still, the IoU metric is steadily growing showing proper improvement with each increase of $T$. As the IoU metrics calculate the semantic context of the environment it is not affected as much by subtle misses from the target. 

Table \ref{Comparison_models} compares all baselines against our model tested on the KITTI-360 dataset. We observe that our config C did best across all metrics. Furthermore, we include the results of config J trained on all datasets and observe that it gives similar results to config C. This suggests that feeding several datasets of various quality and environments does not heavily impact the performance of our model. 

\subsubsection{Comparison in upsampling rate}

Comparing each config in upsampling-only experiments, we observe that config C with the simple 4x upsampling task did best across all metrics. Looking at the scores, it reveals that training on the specific upsampling ratio used during testing yields better results than upsampling with non-matching ratios such as config B and D. Furthermore, the mixed upsampling config A has almost as good performance across all metrics as config C, this shows that the model is generalizing well and might perform better than config C when tested on varied upsampling tasks. 

\subsubsection{Comparison in mask combinations}

Amongst the mixed training task experiments, we observe that config G did best across all metrics for the mixed task training configs. Furthermore, we observe that config G is the only config not trained on the jitter lines. This suggests that the jitter lines did not contribute to the performance of our inpainting DDPM, possibly because of its complexity. Although config G did best across the mixed training models it does not outperform the upsampling only config C. This suggests that although Saharia et al.~\cite{saharia2022palette} showed improvement with multi-task learning in the field of natural images, it does not directly apply to the field of LiDAR. However, our experiments are limited to testing on upsampling only, meaning it is possible that config G would perform better when testing on several tasks at the same time such as upsampling, and noise reduction.

\subsubsection{Multi-dataset learning}
We observe from config J that training on several datasets varying in environments barely affects the overall performance of our inpainting DDPM. This suggests that our inpainting DDPM can generalize well across datasets even when environments and the quality of data vary. Considering the high performance and similarities in setup to the baseline we will use config C as our choice of config for further comparisons unless otherwise specified. 

\subsubsection{Sampling efficiency}
\begin{figure*}[ht!]
    \centering
    \includegraphics[width=\textwidth]{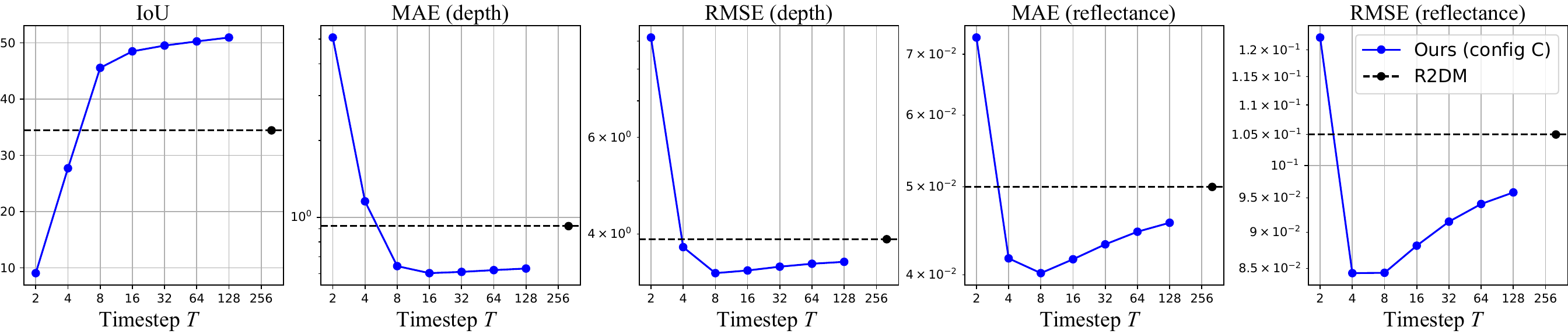}
    \caption[Metric plots]{All metrics comparing our best model configuration C with the baseline R2DM. All metrics but IoU have been log normalized for visual purposes.}
    
    \label{best_model_compare}
\end{figure*}

In Fig. \ref{best_model_compare}, we present a comparative analysis between R2DM and our 4x upsampling model (config C), to describe the trade-off between computational time and performance. Our evaluation includes all timesteps $T$ utilized during the training of our DDPM \{2,4,8,16,32,64,128\}. Despite minor fluctuations in MAE and RMSE observed for reflectance, we generally see a performance increase across all metrics as $T$ increases as expected. Fig. \ref{best_model_compare} illustrates that our inpainting DDPM model outperforms R2DM even for significantly fewer steps. Table \ref{inference_speed} shows the averaged inference speed over three runs for config C and the baseline R2DM~\cite{nakashima2023generative}. We measure our model to be approximately 39 times faster than the R2DM baseline.

\subsubsection{Qualitative results}
\begin{figure*}[ht!]
    \centering
    \scriptsize
    \begin{tabularx}{0.85\hsize}{C|CCCC}
         Input ($16\times1024$) & Nearest Neighbor & Bilinear & Bicubic & LSR~\cite{shan2020simulation} \\
         \includegraphics[width=\hsize]{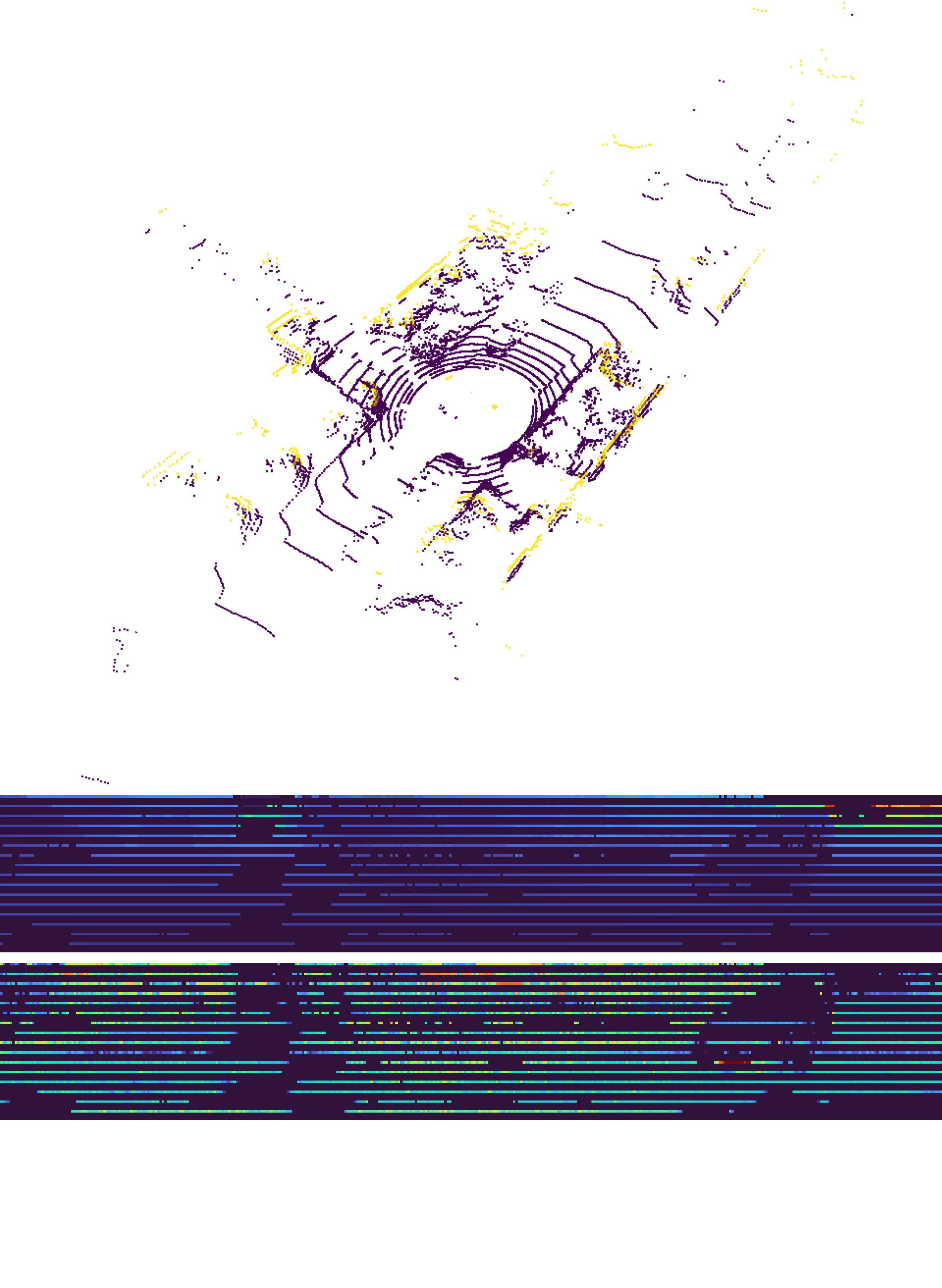} &
         \includegraphics[width=\hsize]{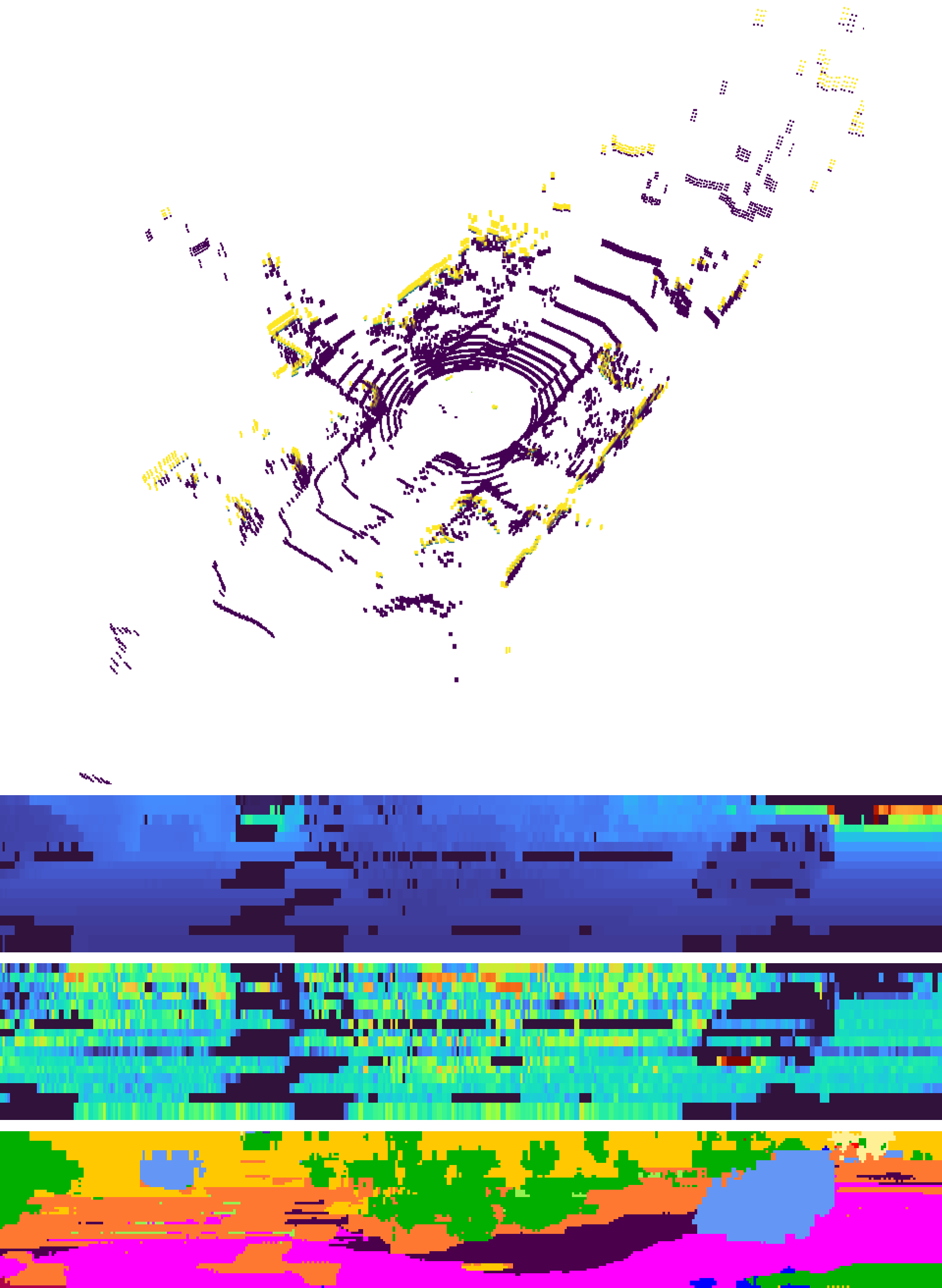} &
         \includegraphics[width=\hsize]{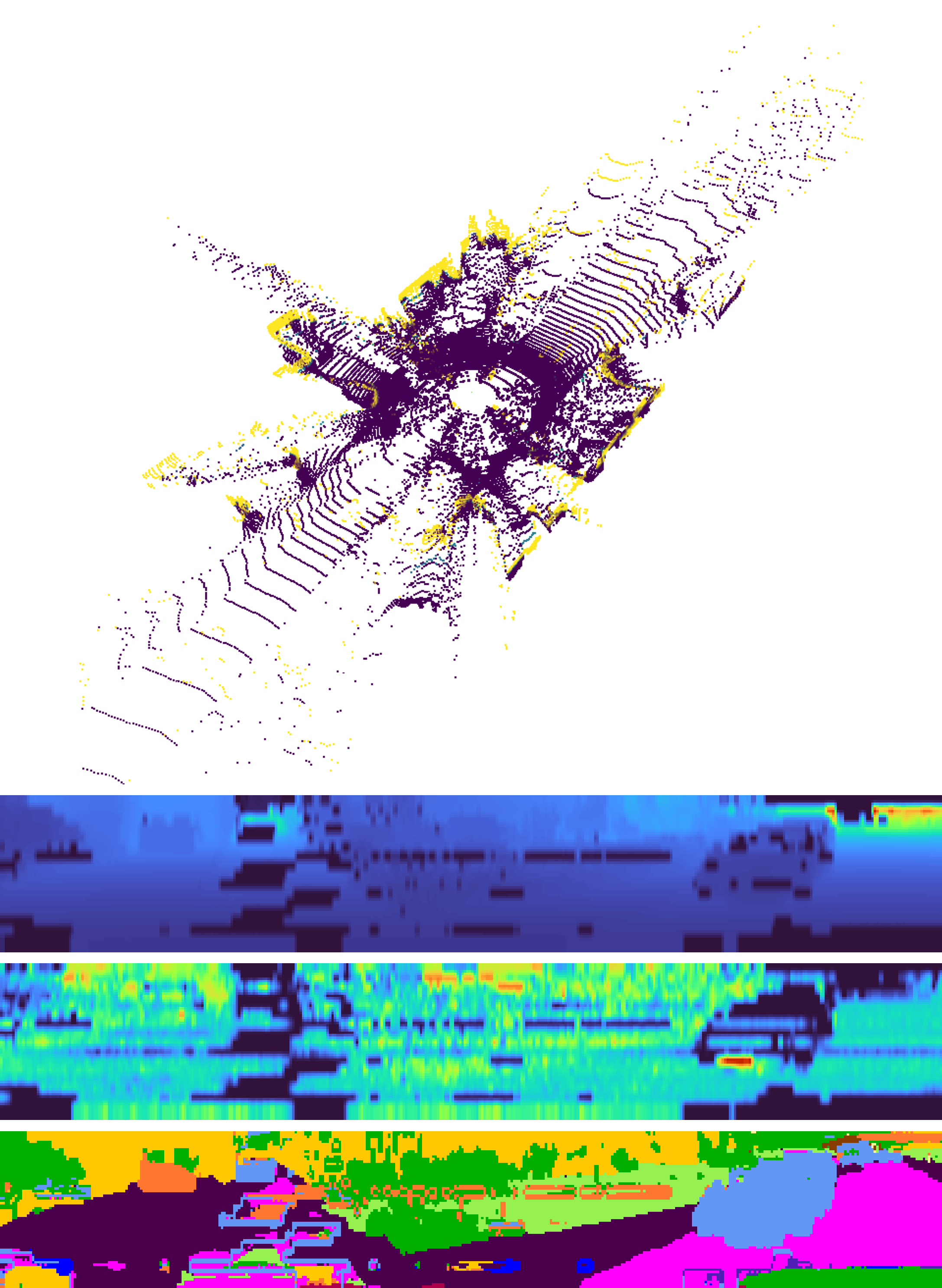} &
         \includegraphics[width=\hsize]{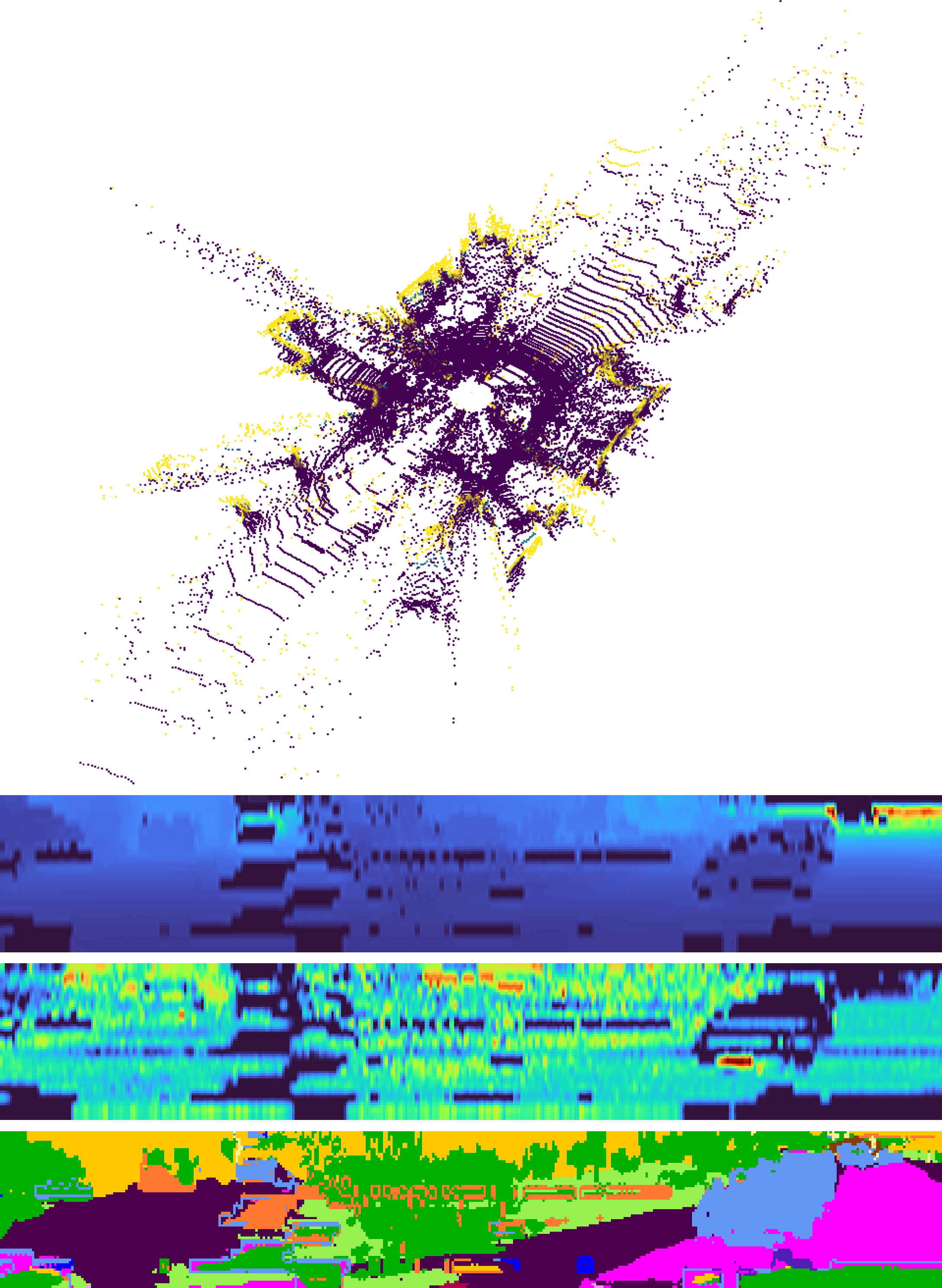} &
         \includegraphics[width=\hsize]{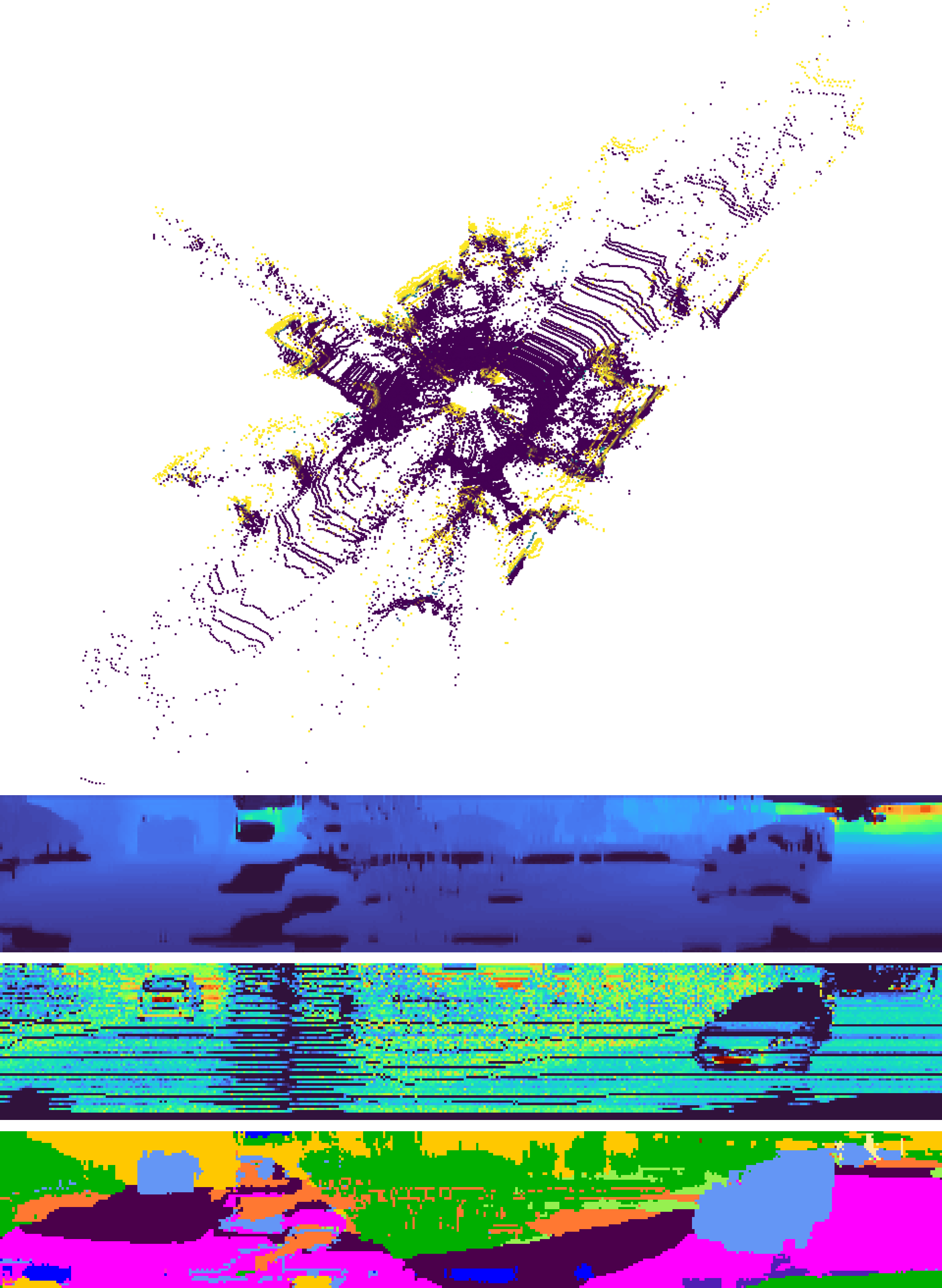} \\
         \\
         Ground Truth ($64\times1024$) & ILN~\cite{kwon2022implicit} & LiDARGen~\cite{zyrianov2022learning} & R2DM~\cite{nakashima2024lidar} & \textbf{Ours} \\
         \includegraphics[width=\hsize]{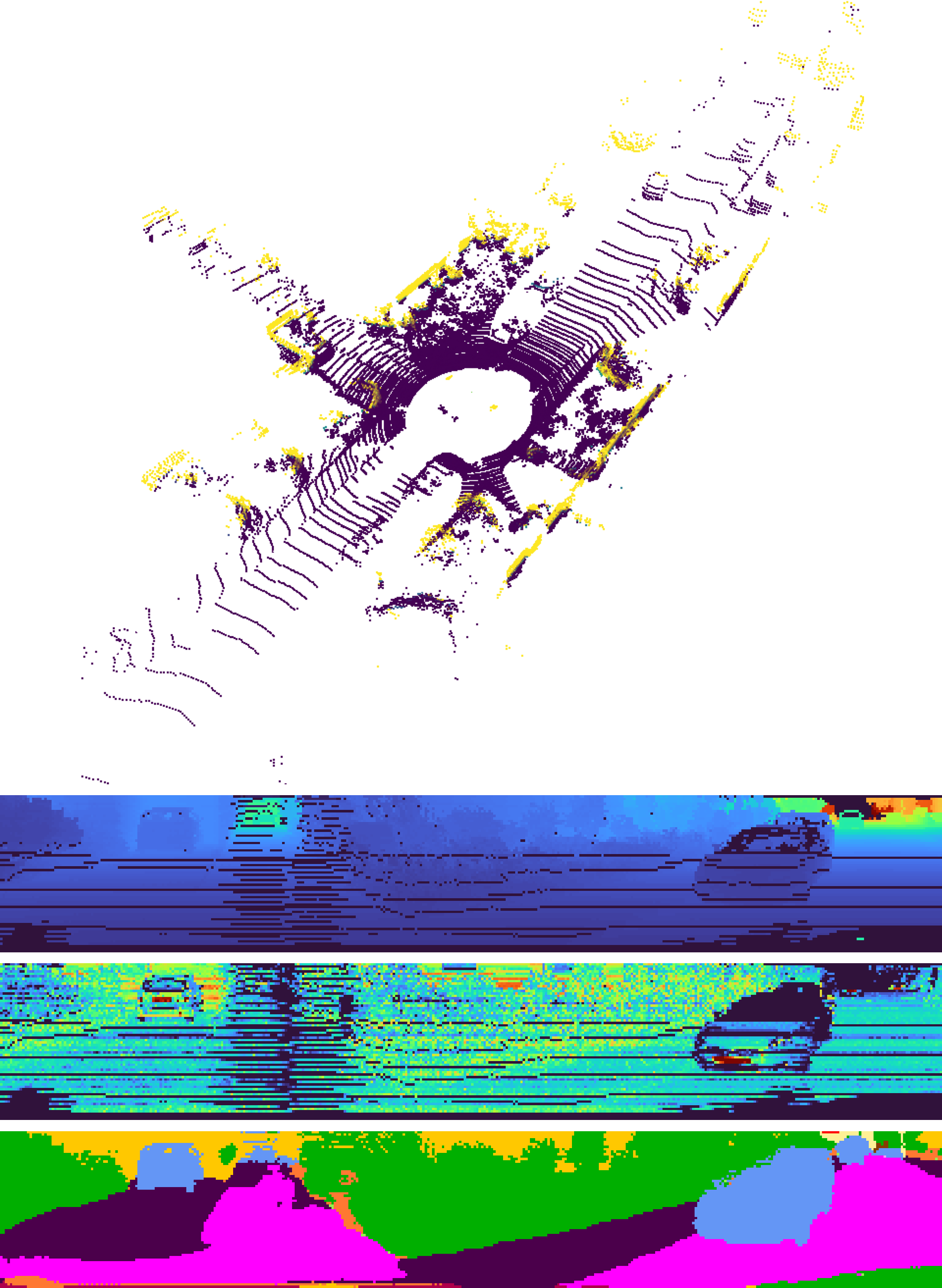} &
         \includegraphics[width=\hsize]{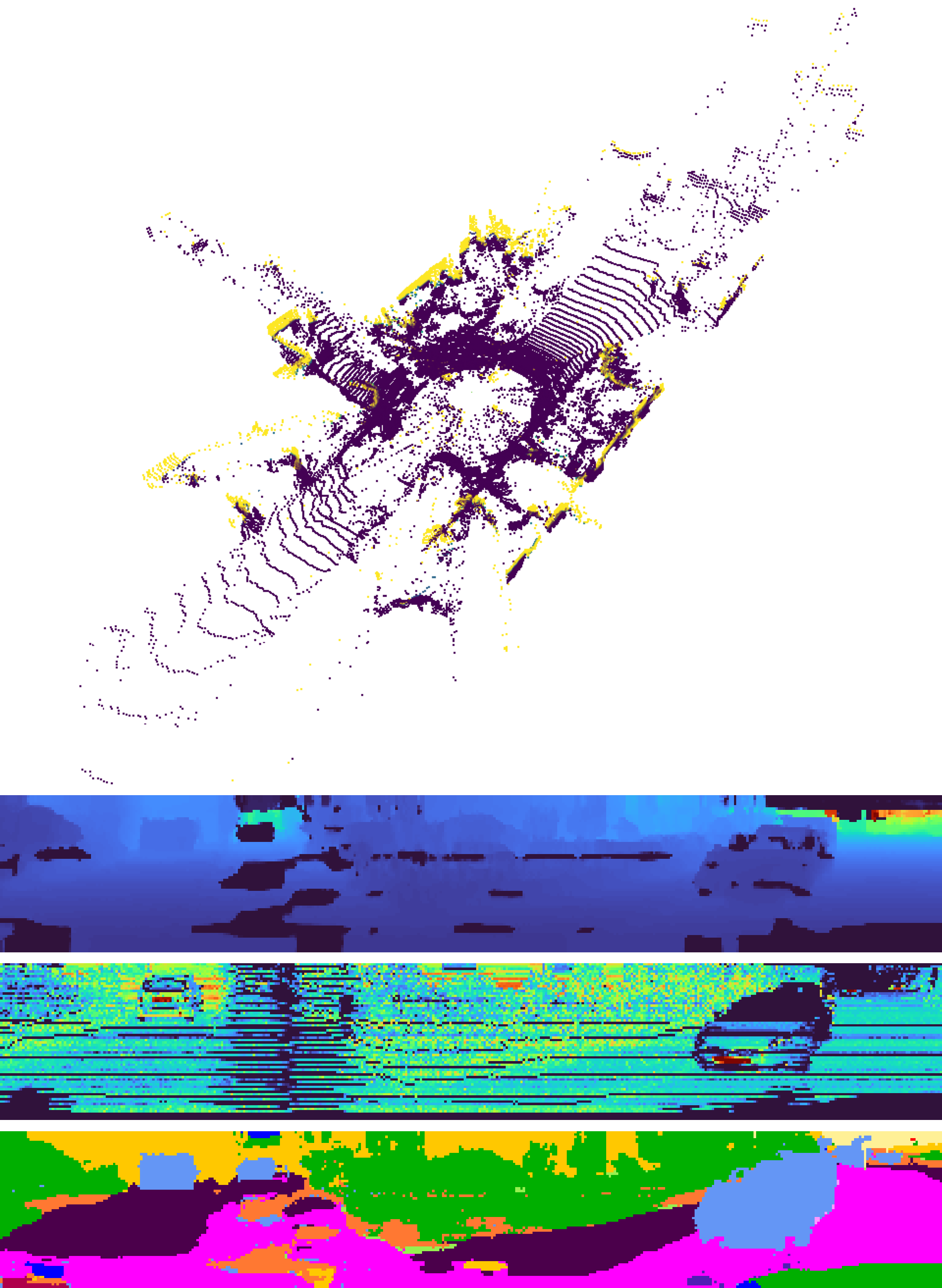} &
         \includegraphics[width=\hsize]{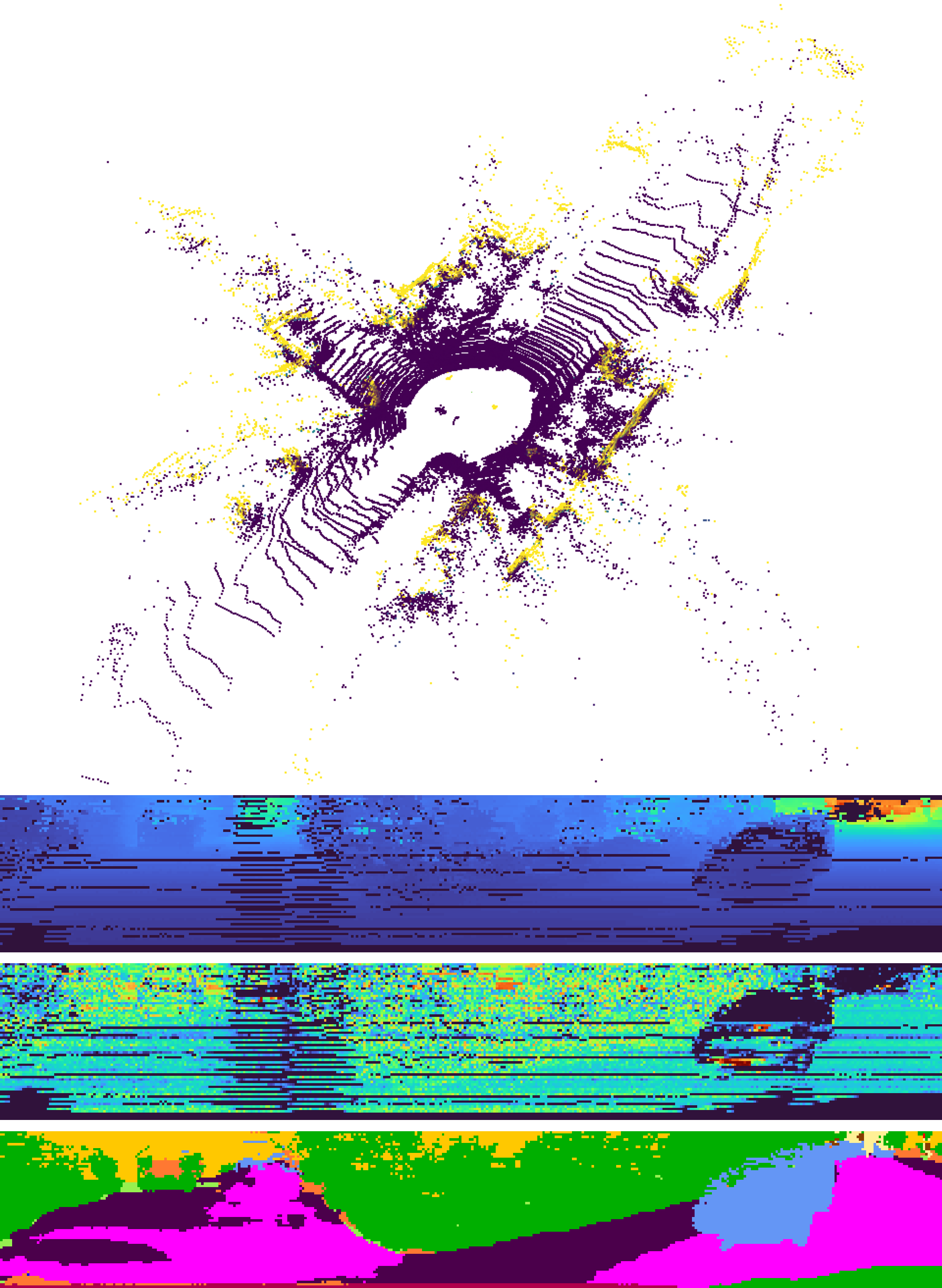} &
         \includegraphics[width=\hsize]{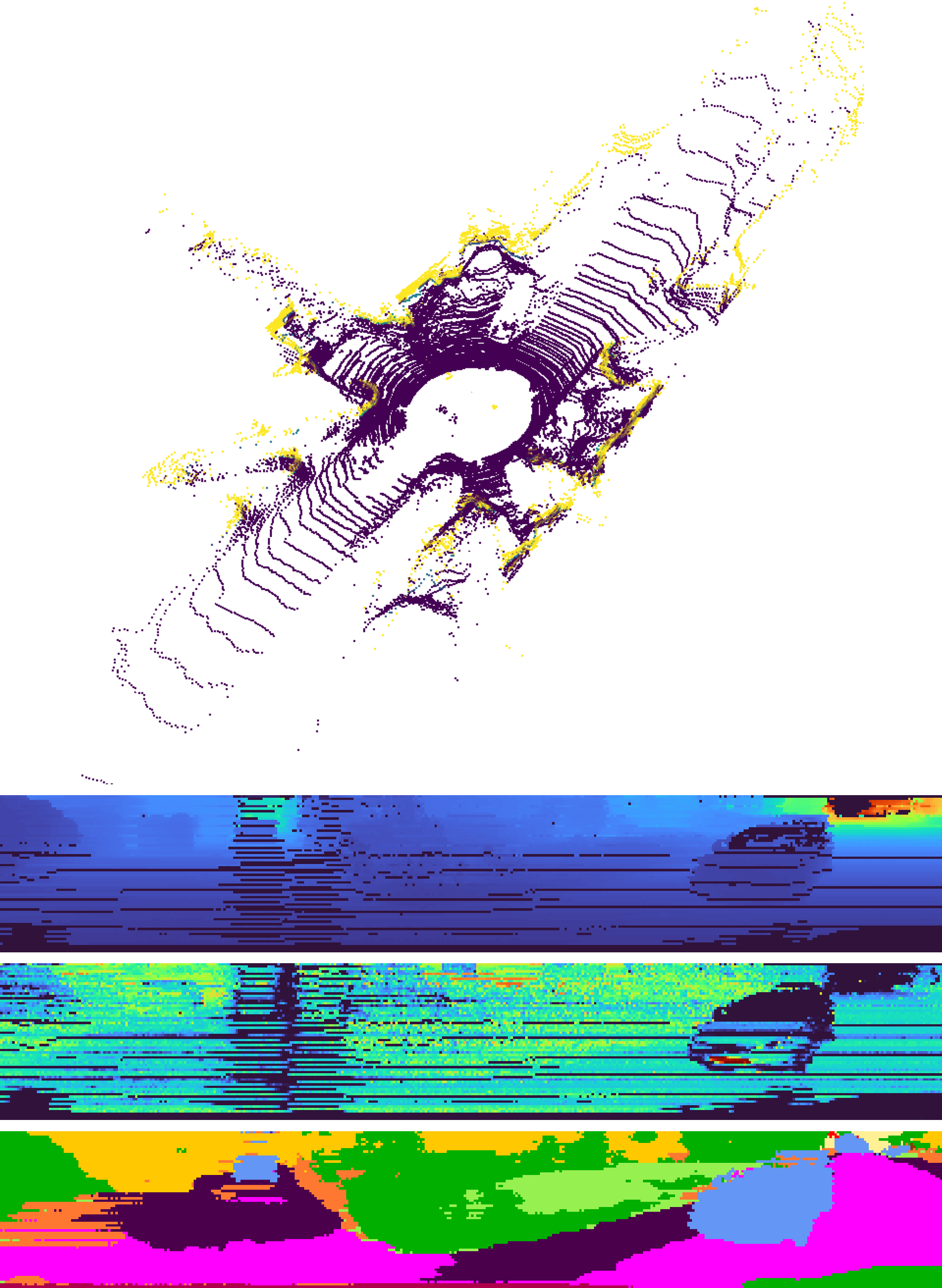} &
         \includegraphics[width=\hsize]{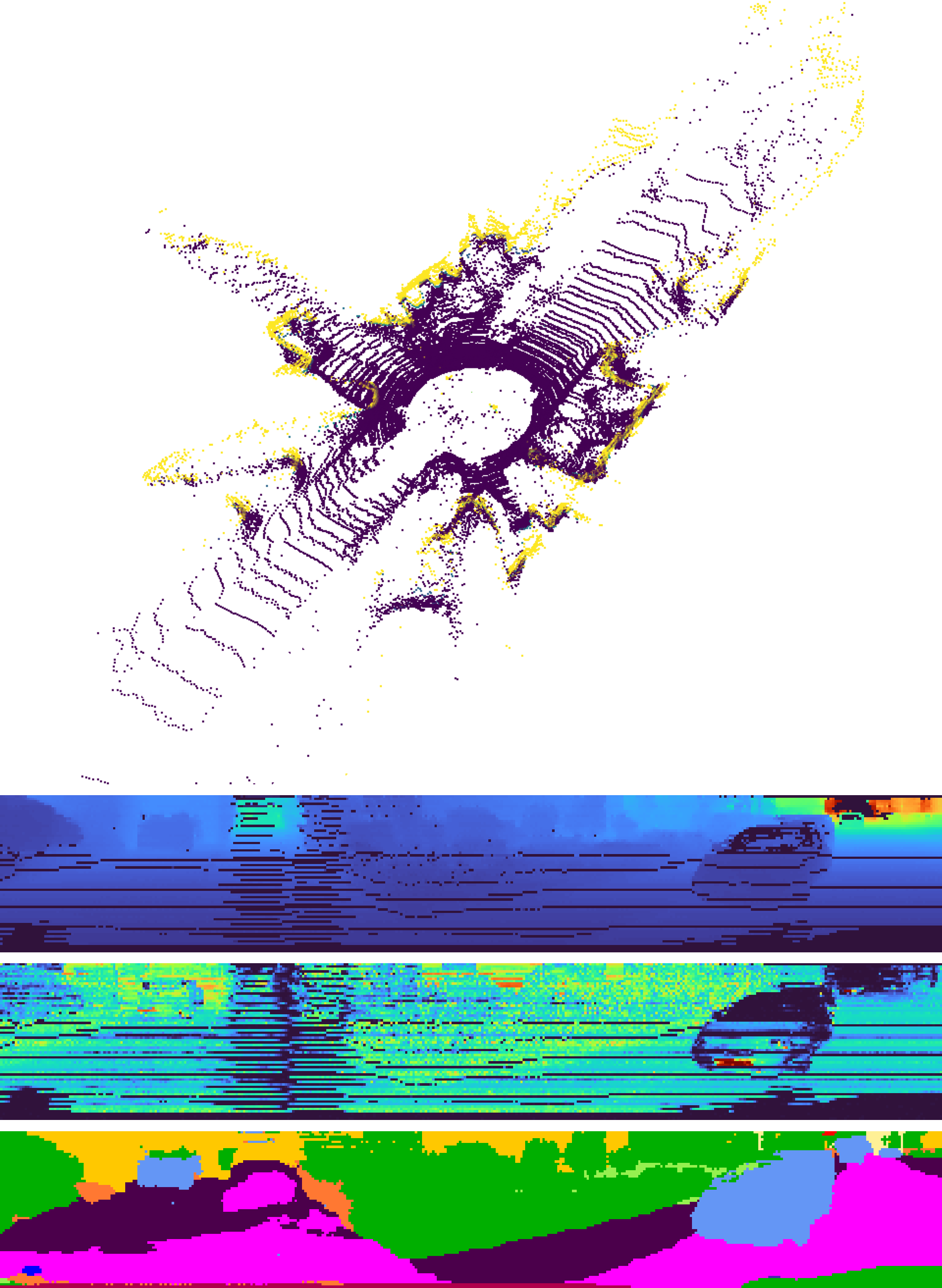} \\
    \end{tabularx}
    \caption[Baseline comparison]{Comparing the results of all baselines with Ours (Config C). The results are reprinted with point clouds on top, followed by range then reflectance images and lastly semantic segmentation results.}
    \label{baselines_comparison}
\end{figure*}

 We observe from the point clouds in Fig. \ref{baselines_comparison} that our method is comparably similar to R2DM. However, when looking at the semantic segmentation we can see that our method is notably more accurate to the ground truth in recreating certain elements compared to R2DM.

\begin{table}[t]
\centering
\caption{Comparison of 4x upsampling on the KITTI-360 test dataset}
\label{Comparison_models}
\begin{tabular}{cl lll}
\toprule
& & \multicolumn{1}{l}{\textbf{Depth}} & \multicolumn{1}{l}{\textbf{Reflectance}} & \\
& \textbf{Method}& \textbf{MAE $\downarrow$}& \textbf{MAE $\downarrow$}& \textbf{IoU\% $\uparrow$}\\
\midrule
\multirow{3}{*}{I}
& Nearest-neighbor& 2.083& 0.106& 18.78\\
& Bilinear& 2.110& 0.101& 18.17\\
& Bicubic& 2.297& 0.108& 18.54\\
\midrule
\multirow{2}{*}{II}
& ILN~\cite{kwon2022implicit}& 2.237& N/A& 21.80\\
& LiDAR-SR~\cite{shan2020simulation}& 2.085& N/A& 21.61\\
\midrule
\multirow{3}{*}{III}
& LiDARGen~\cite{zyrianov2022learning} & 1.551& 0.080& 22.46\\
& R2DM~\cite{nakashima2024lidar} & 0.923& 0.050& 34.44\\
& Ours (config C) & \textbf{0.643}& \textbf{0.040}& \textbf{45.55}\\
\bottomrule
\end{tabular}
\end{table}

\begin{table}[t]
\centering
\caption{Comparison in inference speed}
\label{inference_speed}
\begin{tabularx}{\hsize}{l CCC}
\toprule
\textbf{Method}& \textbf{Timesteps}& \textbf{Time (sec)} & \textbf{FPS}\\
\midrule
R2DM~\cite{nakashima2024lidar} & 320 & 5.373 & 0.19\\
Ours (config C) & \textbf{8} & \textbf{0.137} & \textbf{7.30}\\
\bottomrule
\end{tabularx}
\end{table}

\section{Conclusions}

In this paper, we have proposed a denoising diffusion probabilistic model for an inpainting pipeline for fast upsampling of LiDAR data. Our experiments showed our model was capable of upsampling LiDAR data with high fidelity at 7.30 FPS, which is 39 times faster than the baseline.
We further illustrated the generalization capabilities of our model across dataset-variety regarding environments and quality. 
In addition, we showed that multi-task learning was not necessary for the performance of our upsampling pipeline. 
Future work involve hyperparameter tuning of the model to see if performance can be further improved, and testing with more datasets in various environments such as indoors and adverse weather situations.

\normalem

\bibliographystyle{ieeetr}
\bibliography{root}

\end{document}